\documentclass{article}
\def\arxivpreprint{1}
\usepackage{iclr2027_conference,times}

\usepackage{amsmath,amsfonts,bm}

\def\eqref#1{equation~\ref{#1}}
\def\1{\bm{1}}

\DeclareMathAlphabet{\mathsfit}{\encodingdefault}{\sfdefault}{m}{sl}
\SetMathAlphabet{\mathsfit}{bold}{\encodingdefault}{\sfdefault}{bx}{n}

\usepackage{amsmath,amssymb,amsthm}

\usepackage{algorithm}
\usepackage{algorithmic}

\newtheorem{proposition}{Proposition}[section]
\theoremstyle{remark}
\newtheorem{remark}[proposition]{Remark}
\AtBeginDocument{\crefname{remark}{Remark}{Remarks}}

\usepackage{booktabs}
\usepackage{multirow}
\usepackage{makecell}
\usepackage{graphicx}  
\usepackage{subcaption}
\usepackage{placeins}
\usepackage{wrapfig}

\usepackage{hyperref}
\usepackage[nameinlink,capitalise]{cleveref}
\usepackage{url}
\hypersetup{hidelinks}

\newcommand{\method}{RIDE}

\title{The Teacher Is a Direction, Not a Destination: Extrapolating RL-Induced Representation Residuals in On-Policy Distillation}

\author{\begin{minipage}{\dimexpr\textwidth-2\tabcolsep\relax}
\centering
\textbf{Hao Li}\textsuperscript{1,*}\quad
\textbf{MeiJia Chen}\textsuperscript{2,*}\quad
\textbf{Weijie Ren}\textsuperscript{3,*}\quad
\textbf{Donghan Li}\textsuperscript{4}\\[3pt]
\textbf{Zijun Tian}\textsuperscript{4}\quad
\textbf{Jingchun Huang}\textsuperscript{4}\quad
\textbf{Naibo Wang}\textsuperscript{3,\textdagger}\\[6pt]
{\normalfont\small
\textsuperscript{1}University of Science and Technology of China\\
\textsuperscript{2}Rutgers University\quad
\textsuperscript{3}Zhejiang University\\
\textsuperscript{4}Independent Researcher\\[5pt]
\footnotesize
\begin{tabular}{@{}cc@{}}
\texttt{haoli2101@mail.ustc.edu.cn} &
\texttt{mc2989@rutgers.edu}\\
\texttt{3200101501@zju.edu.cn} &
\texttt{mikeli1997qwe@gmail.com}\\
\texttt{vaynetian@gmail.com} &
\texttt{jh2688@cornell.edu}\\
\multicolumn{2}{c}{\texttt{wangnaibo@zju.edu.cn}}
\end{tabular}\\[4pt]
\textsuperscript{*}Equal contribution.\quad
\textsuperscript{\textdagger}Corresponding author.\endgraf}
\end{minipage}}

\ifdefined\arxivpreprint
  \iclrfinalcopy
\fi
\begin{document}

\maketitle
\ifdefined\arxivpreprint
  \lhead{Preprint}
\fi

\begin{abstract}
On-policy distillation (OPD) trains a student to match the teacher's next-token distributions on the student's own trajectories and has yielded substantial empirical gains. Generalized variants allow the student to surpass the teacher by extrapolating an implicit reward in output space. The language-model head, however, attenuates this change anisotropically: much of the change encoded in the teacher's hidden states reaches the logits at a small fraction of its weight, and the sampled-token log-probability ratios on which output-space extrapolation relies inject noise that the extrapolation amplifies, making training unstable. We observe that reinforcement learning (RL) shifts a model's internal representations relative to its base checkpoint, and that the direction of this shift can be measured at every layer. Motivated by this observation, we propose \method{} (\textbf{R}L-\textbf{I}nduced \textbf{D}irection \textbf{E}xtrapolation), which extrapolates the RL-induced change directly in representation space: at every layer and token position, \method{} computes the residual between the teacher and its pre-RL checkpoint and regresses the student's hidden states toward targets displaced beyond the teacher along this residual. Conditioned on a sampled trajectory, this regression is equivalent to maximizing a linear directional reward defined by the residual under a quadratic penalty centered at the teacher, which makes explicit how the objective moves the student along the RL-induced direction while limiting its deviation from the teacher. Across four base/RL-teacher pairs spanning different scales, architectures, and pre-training lineages, \method{} approaches or exceeds the RL-trained teacher on every pair and is the only method whose mean does so, and it consistently outperforms output-space extrapolation, which degrades the student whenever the teacher is close to its base.
\ifdefined\arxivpreprint
Project page: \url{https://github.com/xixixixixxxx/RIDE}.
\fi
\end{abstract}

\section{Introduction}
\label{sec:introduction}

Distillation transfers capability from a teacher to a student \citep{hinton2015distilling}, and on-policy distillation (OPD) has become a standard stage of language-model post-training \citep{yang2025qwen3,xiao2026opd}. In OPD the student samples its own trajectories and receives per-token supervision from the teacher on the contexts it actually visits, which reduces exposure bias and provides a dense learning signal \citep{agarwal2024onpolicy,gu2024minillm,lu2025onpolicy}. A common use is to transfer the gains of a reinforcement-learning (RL) teacher to a student that shares its initialization, for example when merging domain experts or distilling an expensive RL run back into its base model \citep{xiao2026opd,yang2026learning}.

\begin{figure*}[t]
\centering
\includegraphics[width=\textwidth]{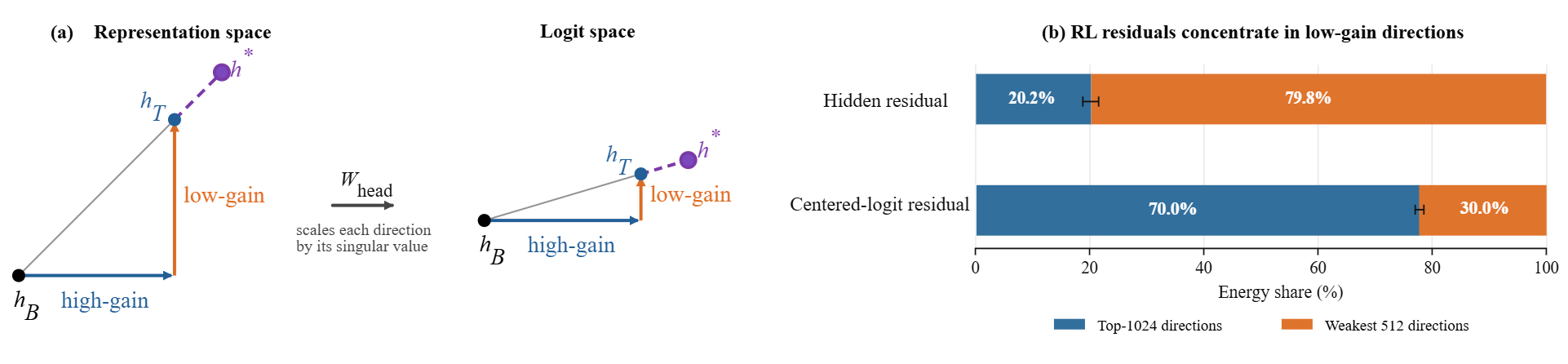}
\caption{\textbf{The language-model head attenuates RL-induced representation changes unevenly.} (a) Residual extrapolation before and after projection through the head. (b) Hidden-state and centered-logit energy shares of the teacher-to-base residual across the singular directions of the head. Whiskers show prompt-bootstrap $95\%$ confidence intervals.}
\label{fig:motivation}
\end{figure*}

Standard OPD treats the teacher as the endpoint of learning, yet students need not stop there: the teacher-to-reference log-probability ratio acts as an implicit reward \citep{rafailov2023dpo,yuan2024free,cui2025process}, up-weighting it relative to the KL constraint yields students that outperform the teacher \citep{yang2026learning}, and related methods extrapolate displacements in parameter or logit space \citep{li2026rise,zheng2025expo}. Because RL updates are small, structured, and consistent across runs \citep{mukherjee2025subnetworks,zhu2025path,shenfeld2026razor}, an RL-trained teacher supplies a direction as well as a destination, and a student that starts from the same checkpoint can continue along it.

Extrapolating in output space, however, measures this direction after the head has attenuated most of it. The language-model head reweights representation changes according to its anisotropic singular spectrum \citep{yang2018softmax,yang2026oprd}: in \cref{fig:motivation}, its $512$ weakest directions carry $79.8\%$ of the teacher-to-base residual's hidden-state energy but only $30.0\%$ of its centered-logit energy, so output-space objectives supervise much of the residual only weakly and leave earlier layers unconstrained. The sampled-token log-ratio is moreover length-biased and prone to overoptimization \citep{park2024length,rafailov2024overopt}, and scaling it by a global coefficient amplifies extreme rewards and destabilizes training \citep{yang2026learning,sun2026reopd,li2026cliff}. Hidden states retain the change before the head: feature-level distillation already uses them as supervision \citep{romero2015fitnets,sun2019patient,jiao2020tinybert}, its on-policy form OPRD matches student and teacher states layerwise with a deterministic per-rollout gradient \citep{yang2026oprd}, and hidden-state directions encode behaviors that can be steered additively \citep{park2024linear,turner2023steering,zou2023repe,arditi2024refusal}, which makes residual extrapolation a natural operation in this space.

We therefore propose \emph{\method{}} (\textbf{R}L-\textbf{I}nduced \textbf{D}irection \textbf{E}xtrapolation). With the student initialized at the teacher's pre-RL checkpoint, the three models share a representation space. On each student-generated context, \method{} evaluates the frozen teacher and the pre-RL checkpoint on identical tokens, takes their layerwise hidden-state difference as the RL-induced residual, places each target beyond the teacher along this residual with a single coefficient, and regresses the student's hidden states toward it. Conditioned on a rollout, this regression maximizes a linear directional reward under a quadratic penalty around the teacher, and under a shared linear head the output-space extrapolation target of \citet{yang2026learning} is the head-projected image of the \method{} target; \method{} thus retains the reward-extrapolation view while replacing the sampled-token log-ratio with a deterministic hidden-state gradient.

On four base/RL-teacher pairs (R1-Distill-1.5B, Qwen3-4B, Llama-3.2-3B, and Phi-4-mini), which span different scales, model families, and vocabularies, \method{} attains higher mean Avg@16 on AIME24, AIME25, and AIMO than OPRD and the output-space baselines on every pair and approaches or exceeds the RL-trained teacher itself, whereas output-space extrapolation falls below its teacher on every pair.
 
Our contributions are as follows.
\begin{itemize}
    \item We identify the RL-induced representation residual, the layerwise difference between the teacher and its pre-RL checkpoint on identical on-policy contexts, as a measure of how RL changes the model's computation. Both checkpoints are available in the same-initialization distillation setting.
    \item We propose \method{}, which extrapolates hidden-state targets along this residual with a single coefficient that recovers OPRD at $\lambda=1$, and we interpret the objective as reward maximization under a quadratic penalty, the representation-space counterpart of output-space reward extrapolation.
    \item We show that \method{} approaches or exceeds its RL-trained teacher on all four base/teacher pairs while consistently outperforming OPRD and output-space extrapolation, and we characterize its learning signal through analyses of head attenuation and conditional variance.
\end{itemize}

\section{Related Work}
\label{sec:related}

\paragraph{On-policy distillation.}
Unlike sequence-level imitation \citep{hinton2015distilling,kim2016sequence}, on-policy distillation samples from the student and supervises the prefixes it visits \citep{agarwal2024onpolicy,gu2024minillm,ko2024distillm}; its sampled-token form is now a standard post-training stage \citep{lu2025onpolicy,yang2025qwen3,xiao2026opd}. Recent variants remove the need for teacher logits \citep{ye2025blackbox}, reformulate the target for stability \citep{jang2026stable,jin2026entropy}, analyze failure modes \citep{fu2026revisiting,li2026rethinking}, or use the model itself, given privileged information, as its own teacher \citep{zhao2026opsd,hubotter2026sdpo}. All of these approaches supervise outputs rather than intermediate states.

\paragraph{Learning beyond the teacher.}
Viewing OPD as KL-constrained RL with an implicit log-ratio reward \citep{rafailov2023dpo,yuan2024free,cui2025process}, generalized OPD up-weights this reward and obtains students that exceed the teacher \citep{yang2026learning}. A global coefficient, however, drives the student toward extreme reward peaks \citep{sun2026reopd} and can collapse near-deterministic outputs \citep{li2026cliff}, echoing the length bias and overoptimization of log-ratio rewards \citep{park2024length,rafailov2024overopt}. Other methods extrapolate in parameter or logit space, either along a policy's own RL trajectory \citep{li2026rise} or along the difference between a fine-tuned model and its initialization \citep{ilharco2023task,zheng2025expo}. \method{} shares the premise that the RL-induced change is a usable direction, but measures it on hidden states, at every layer, before the head, and on the student's own contexts.

\paragraph{Representation-level distillation.}
Intermediate representations have long served as distillation targets \citep{romero2015fitnets,zagoruyko2017attention,tian2020crd,sun2019patient,jiao2020tinybert,wang2020minilm}. OPRD applies layerwise hidden-state regression on policy, obtaining a deterministic per-rollout gradient and supervision that output-space losses cannot see through the head \citep{yang2026oprd}; PHF instead matches the directions and geometry of hidden-state transitions \citep{li2026phf}. \method{} uses the residual from the pre-RL checkpoint to the teacher to place targets beyond the teacher, drawing on the approximately linear structure of behavioral directions in hidden space \citep{park2024linear,turner2023steering,zou2023repe,arditi2024refusal} and on the structured, consistent nature of RL updates \citep{mukherjee2025subnetworks,zhu2025path,shenfeld2026razor}. Latent memory and internal-state interventions have also been studied in medical vision-language models \citep{zhu2026medsynapse,zhu2026craft}.

\section{Preliminaries}
\label{sec:prelim}

\paragraph{Notation.}
Let $\pi_\theta$ be the trainable student and $\pi_T$ the frozen teacher. Both are decoder-only transformers with $L$ layers, hidden dimension $d$, a shared vocabulary $\mathcal{V}$, and a language-model head $W_{\mathrm{head}}\in\mathbb{R}^{|\mathcal{V}|\times d}$. For a prompt $x\sim\mathcal{D}_x$, the student samples $\hat y=(\hat y_1,\dots,\hat y_T)\sim\pi_\theta(\cdot\mid x)$. We evaluate all models on the same student-generated prefix $(x,\hat y_{<t})$ and write $\pi(\cdot\mid x,\hat y_{<t})$ for a model's next-token distribution and $h^{(l)}_{\pi,t}\in\mathbb{R}^d$ for its layer-$l$ residual-stream state at position $t$.

\paragraph{On-policy distillation.}
Classical knowledge distillation trains the student on teacher outputs or teacher samples \citep{hinton2015distilling,kim2016sequence}. On-policy distillation (OPD) instead samples responses from the student and minimizes the reverse KL divergence to the teacher on the prefixes it visits \citep{agarwal2024onpolicy,gu2024minillm}; by the chain rule of the KL divergence,
\begin{equation}
\label{eq:opd}
\mathcal{L}_{\mathrm{OPD}}(\theta)=\mathbb{E}_{x,\;\hat y\sim\pi_\theta}\Big[\sum_{t=1}^{T}D_{\mathrm{KL}}\big(\pi_\theta(\cdot\mid x,\hat y_{<t})\,\big\|\,\pi_T(\cdot\mid x,\hat y_{<t})\big)\Big].
\end{equation}
Estimating each local KL from the sampled token gives the commonly used tokenwise update
\begin{equation}
\label{eq:opd-grad}
g_{\mathrm{OPD}}^{\mathrm{token}}(\theta)=\mathbb{E}_{x,\;\hat y\sim\pi_\theta}\Big[\sum_{t=1}^{T}r_t\,\nabla_\theta\log\pi_\theta(\hat y_t\mid x,\hat y_{<t})\Big],
\qquad
r_t=\log\frac{\pi_\theta(\hat y_t\mid x,\hat y_{<t})}{\pi_T(\hat y_t\mid x,\hat y_{<t})},
\end{equation}
in which $-r_t$ acts as a dense per-token advantage \citep{lu2025onpolicy,xiao2026opd}. Relative to any reference policy $\pi_{\mathrm{ref}}$, the same objective is KL-regularized RL with the implicit reward $\log\pi_T-\log\pi_{\mathrm{ref}}$ \citep{rafailov2023dpo,yang2026learning}; output-space extrapolation scales this reward (\cref{app:opd}).

\paragraph{On-policy supervision as target matching.}
Let $s_{\pi,t}$ be a model quantity at position $t$, $\tau_t$ its target, and $\ell$ a discrepancy measure:
\begin{equation}
\label{eq:template}
\mathcal{L}(\theta)=\mathbb{E}_{x\sim\mathcal{D}_x,\;\hat y\sim\pi_\theta(\cdot\mid x)}\Big[\frac{1}{T}\sum_{t=1}^{T}\ell\big(s_{\theta,t},\;\mathrm{sg}(\tau_t)\big)\Big],
\end{equation}
where $\mathrm{sg}$ stops gradients through the target. Output-space teacher matching takes $s_{\pi,t}$ to be the next-token distribution and $\ell$ the reverse KL divergence. Representation matching takes $s_{\pi,t}=h^{(l)}_{\pi,t}$ over a set of layers $l\in\mathcal{L}_{\mathrm{layer}}$ and over positions selected by a mask $m_t\in\{0,1\}$, with the dimension-normalized squared error
\begin{equation}
\label{eq:rep-loss}
\ell\big(h^{(l)}_{\theta,t},\tau^{(l)}_t\big)=\frac{1}{d}\big\|h^{(l)}_{\theta,t}-\tau^{(l)}_t\big\|_2^2,
\end{equation}
which is on-policy feature distillation \citep{romero2015fitnets,jiao2020tinybert,yang2026oprd}; conditioned on a rollout, its gradient is deterministic, unlike that of \cref{eq:opd-grad}. Both instances set $\tau_t=s_{T,t}$, so the teacher is the endpoint of learning.

\paragraph{Same-initialization setting.}
We study a teacher obtained by RL from a base checkpoint, $\pi_T=\mathrm{RL}(\pi_B)$, and a student initialized at that checkpoint, $\pi_{\theta_0}=\pi_B$. This setting arises when an RL result is distilled back into its base model or when RL-trained experts that share a base are merged \citep{xiao2026opd,yang2026learning}. The three models share an architecture, a tokenizer, and a representational origin, so on a common prefix the difference $s_{T,t}-s_{B,t}$ measures what RL changed on that context: in output space it is the implicit reward $\log\pi_T-\log\pi_B$ \citep{rafailov2023dpo}; in representation space it is a vector at every layer, taken before the head.

\section{Method}
\label{sec:method}

\begin{figure*}[t]
\centering
\includegraphics[width=\textwidth]{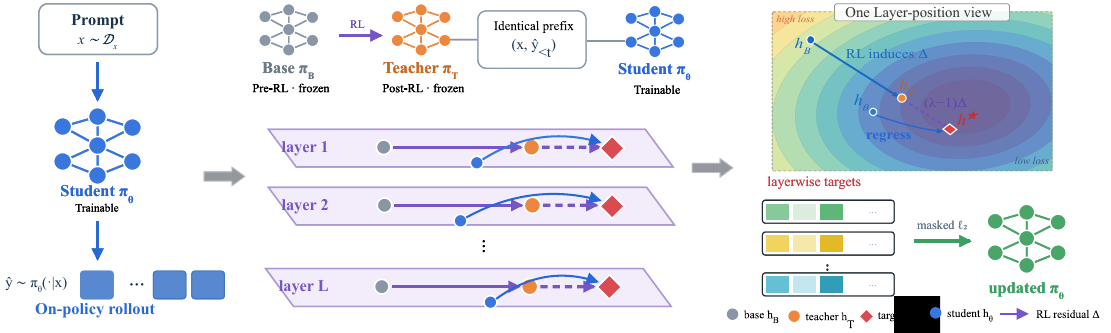}
\caption{\textbf{Overview of \method{}.} On student-generated prefixes, the frozen base and teacher define layerwise residuals. \method{} extrapolates hidden-state targets along these residuals and trains the student by masked regression toward them. Right: target geometry at a single layer and position.}
\label{fig:method}
\end{figure*}

\method{} rests on the premise that an RL-trained teacher specifies a direction in addition to a destination: the teacher's displacement from $\pi_B$ is what RL contributed, and a student initialized at $\pi_B$ can continue along it (\cref{fig:method}). We state the construction in the target-regression template (\cref{sec:method-principle}), instantiate it on hidden states (\cref{sec:method-residual}), justify this choice of space (\cref{sec:method-why}), and interpret the objective as reward maximization (\cref{sec:method-analysis}; implementation in \cref{sec:method-impl}).

\subsection{From Teacher Matching to Residual Extrapolation}
\label{sec:method-principle}

Within \cref{eq:template}, teacher matching is the choice $\tau_t=s_{T,t}$. We instead displace the target from the teacher along the RL-induced change:
\begin{equation}
\label{eq:principle}
\tau_t(\lambda)\;=\;s_{T,t}+(\lambda-1)\big(s_{T,t}-s_{B,t}\big)\;=\;\lambda\,s_{T,t}+(1-\lambda)\,s_{B,t},
\qquad \lambda\ge 0 .
\end{equation}
At $\lambda=1$ the target is the teacher and \cref{eq:template} is recovered; for $\lambda\in(0,1)$ it lies between base and teacher; for $\lambda>1$ it lies beyond the teacher, so the student continues the change that RL initiated. Because $s_{T,t}-s_{B,t}$ is recomputed for every rollout and prefix, the direction is not a fixed offset but the RL-induced change on the contexts the student visits. On log-probabilities, \cref{eq:principle} yields the affine combination $\lambda\log\pi_T+(1-\lambda)\log\pi_B$ that underlies output-space reward extrapolation \citep{yang2026learning}. We argue below that this measures the displacement only after the head has attenuated most of it, and instead instantiate \cref{eq:principle} on hidden states.

\subsection{RL-Induced Representation Residual and the \method{} Objective}
\label{sec:method-residual}

Given a student rollout, the frozen models $\pi_T$ and $\pi_B$ are evaluated on the same prefix $(x,\hat y_{<t})$, and the \emph{RL-induced residual} at layer $l$ and position $t$ is $\Delta^{(l)}_t=h^{(l)}_{T,t}-h^{(l)}_{B,t}$. Because both models process the same tokens, $\Delta^{(l)}_t$ isolates what RL changed in the processing of that context rather than the effect of a different trajectory; it lives in the student's hidden-state space, is defined at every layer, and is computed deterministically from two forward passes rather than sampled.

Applying \cref{eq:principle} to hidden states yields the \method{} target,
\begin{equation}
\label{eq:target}
h^{\star(l)}_{t}\;=\;h^{(l)}_{T,t}+(\lambda-1)\,\Delta^{(l)}_t\;=\;\lambda\,h^{(l)}_{T,t}+(1-\lambda)\,h^{(l)}_{B,t},
\end{equation}
and the training objective is the representation-space regression of \cref{eq:rep-loss} toward this target,
\begin{equation}
\label{eq:ride}
\mathcal{L}_{\lambda}(\theta)=\mathbb{E}_{x,\;\hat y\sim\pi_\theta}\Big[\frac{1}{|\mathcal{L}_{\mathrm{layer}}|}\sum_{l\in\mathcal{L}_{\mathrm{layer}}}\frac{1}{M}\sum_{t=1}^{T}m_t\,\frac{1}{d}\big\|h^{(l)}_{\theta,t}-\mathrm{sg}\big(h^{\star(l)}_{t}\big)\big\|_2^2\Big],
\end{equation}
where $M=\sum_t m_t$. Unlike its log-probability counterpart, \cref{eq:target} involves no partition function. We supervise all $L$ layers and the final $k$ response positions, where student--teacher disagreement concentrates \citep{yang2026oprd}. Only the target differs from representation matching, so $\lambda=1$ recovers OPRD exactly and any difference between the two methods is due solely to the residual term. Because \cref{eq:rep-loss} is unnormalized, the loss scale grows with $\lambda$; we keep this form so that $\lambda=1$ reduces exactly to OPRD and rescale the loss coefficient instead (\cref{rem:scale}).

\subsection{Why the Displacement Should Be Measured in Hidden States}
\label{sec:method-why}

Two arguments favor hidden states over log-probabilities in \cref{eq:principle}: how much of the residual survives the head, and how estimation noise behaves when the target is displaced beyond the teacher.

\paragraph{The head attenuates the residual anisotropically.}
Let $\tilde h$ denote the representation fed to the head and $z=W_{\mathrm{head}}\tilde h$ the logits. Since the head is linear, displacing its input along the residual displaces the logits along the corresponding logit residual,
\begin{equation}
\label{eq:logit-image}
W_{\mathrm{head}}\big(\tilde h_T+(\lambda-1)\tilde\Delta\big)=z_T+(\lambda-1)(z_T-z_B)=\lambda z_T+(1-\lambda)z_B .
\end{equation}
Up to the log-partition constant removed by the softmax, the right-hand side is the log-probability target of \cref{eq:principle}; the output-space target is thus the image of the \method{} target under the head. The converse fails for two reasons. First, a logit target constrains $\tilde h$ only weakly along the directions that $W_{\mathrm{head}}$ amplifies least, and \cref{sec:exp-mechanism} shows that the RL-induced residual concentrates in precisely these directions, so they reach an output-space loss at a small fraction of their weight. Second, a logit target imposes no constraint on layers below the final one, whereas \cref{eq:target} specifies the full vector at every supervised layer. Output-space reward extrapolation is therefore the head projection of representation extrapolation, which extrapolates the residual at full weight in every direction; the assumptions behind \cref{eq:logit-image} and the head-drift term are discussed in \cref{app:head-diagnostics}.

\paragraph{Extrapolation amplifies output-space noise but not hidden-state noise.}
Fix a rollout and a position $t$; write $p=\pi_\theta(\cdot\mid x,\hat y_{<t})$, $q=\pi_T(\cdot\mid x,\hat y_{<t})$, and $b=\pi_B(\cdot\mid x,\hat y_{<t})$, and let $v\sim p$ be the sampled token. With $\ell(v)=\log p(v)-\log q(v)$ and $\rho(v)=\log q(v)-\log b(v)$, output-space extrapolation uses the advantage $A_\lambda(v)=\ell(v)-(\lambda-1)\rho(v)$ in the tokenwise update of \cref{eq:opd-grad} (\cref{app:opd}); both components are evaluated at a single sampled token.

\begin{proposition}[Conditional variance under extrapolation]
\label{prop:variance}
Conditioned on the prefix, the variance of the output-space advantage over the sampled token satisfies
\begin{equation}
\label{eq:var}
\mathrm{Var}_{v\sim p}\big[A_\lambda(v)\big]=\mathrm{Var}_{v\sim p}[\ell(v)]+(\lambda-1)^2\,\mathrm{Var}_{v\sim p}[\rho(v)]-2(\lambda-1)\,\mathrm{Cov}_{v\sim p}[\ell(v),\rho(v)] .
\end{equation}
The second term does not vanish as $p\to q$ unless $\log q-\log b$ is constant on the support of $q$. Conditioned on the same prefix, the gradient of the \method{} objective in \cref{eq:ride} is deterministic for every $\lambda$, so its conditional variance is zero.
\end{proposition}

The proof is in \cref{app:proofs}. In output space the extrapolated component is a sampled estimate of $\rho$ whose noise is scaled by $(\lambda-1)^2$ however close the student is to the teacher, which accounts for the instability reported at large scaling factors \citep{yang2026learning,sun2026reopd,li2026cliff}; in hidden-state space the residual is computed rather than sampled, so $\lambda$ changes only the destination.

\subsection{Optimization Interpretation}
\label{sec:method-analysis}

The regression in \cref{eq:ride} can be read as reward maximization under a proximity constraint. Fix a rollout, a supervised layer, and a position, and write $h=h^{(l)}_{\theta,t}$, $h_T=h^{(l)}_{T,t}$, and $\Delta=\Delta^{(l)}_t$.

\begin{proposition}[Reward-and-penalty form]
\label{prop:reward}
Conditioned on the rollout, minimizing $\mathcal{L}_\lambda$ with respect to the student's hidden state at each supervised layer and position is equivalent, up to a positive multiplicative factor and an additive constant, to maximizing
\begin{equation}
\label{eq:reward-form}
(\lambda-1)\,r(h)\;-\;\tfrac{1}{2}\big\|h-h_T\big\|_2^2,
\end{equation}
where $r(h)=\langle h-h_T,\Delta\rangle$. The maximizer is $h^\star=h_T+(\lambda-1)\Delta$, and $\nabla_h\mathcal{L}_\lambda=\tfrac{2}{d}(h-h^\star)$.
\end{proposition}

The reward $r$ is the inner product of the student's displacement from the teacher with the RL-induced residual, the penalty is quadratic in the same displacement, and $\lambda-1$ weights one against the other; at $\lambda=1$ the reward weight vanishes and OPRD is recovered. This is the representation-space counterpart of the KL-constrained RL view of output-space extrapolation \citep{yang2026learning}; see \cref{app:reward} for the derivation.

\section{Experiments}
\label{sec:experiments}
\subsection{Experimental Setup}
\label{sec:exp-setup}

\paragraph{Models.}
We use four base/RL-teacher pairs; in each, the student is initialized from the base checkpoint (\cref{sec:prelim}). The R1-Distill-1.5B pair consists of DeepSeek-R1-Distill-Qwen-1.5B and JustRL-DeepSeek-1.5B \citep{guo2025deepseekr1,he2025justrl}. The remaining teachers, Just-Qwen3-4B, Just-Llama-3.2-3B, and Just-Phi-4-mini, are obtained by applying the same RL recipe \citep{he2025justrl} to Qwen3-4B \citep{yang2025qwen3}, Llama-3.2-3B \citep{grattafiori2024llama3}, and Phi-4-mini \citep{abouelenin2025phi4mini}. Within each pair, the base, teacher, and student share the architecture, tokenizer, and language-model head, which permits direct hidden-state comparison; across pairs, depth, width, vocabulary, and pre-training lineage all differ.

\paragraph{Training data and protocol.}
Prompts are drawn from DAPO-Math-17K \citep{yu2025dapo}. Each method samples rollouts from its own current student policy, and the frozen teacher and, where needed, the base checkpoint are evaluated on those prefixes. All runs share the prompt dataset, sampling settings, optimizer schedule, and a budget of $500$ training steps. Unless stated otherwise, \cref{sec:exp-lambda,sec:exp-mechanism,sec:exp-ablation} use the R1-Distill-1.5B pair, and \method{} supervises all $L$ layers and the last $k=2{,}000$ response positions with $\lambda=1.25$ and loss scaling by $\lambda^{-2}$ (\cref{rem:scale}). Hyperparameters and hardware are listed in \cref{sec:method-impl}.

\paragraph{Baselines.}
We compare against sampled-token OPD (top-1) and top-16 OPD for output-space teacher matching \citep{agarwal2024onpolicy,lu2025onpolicy}; OPRD for representation-space matching, which coincides with \method{} at $\lambda=1$ \citep{yang2026oprd}; and ExOPD for output-space extrapolation, which uses the same $\lambda$ and pre-RL checkpoint \citep{yang2026learning}. The frozen teacher and the untouched student serve as reference points. Scaling coefficients are swept over the same grid as for \method{}. We omit weight-space extrapolation because ExOPD was stronger in the same setting \citep{zheng2025expo,yang2026learning}.

\paragraph{Evaluation.}
We report Avg@16 on AIME 2024, AIME 2025, and AIMO (AMC 2022--2023) \citep{aimo2024aime,opencompass2025aime,aimo2024amc}. For each training run, we average correctness over $16$ sampled responses per problem and then over problems within each benchmark; Avg. is the unweighted mean of the three benchmark scores. We use three independent training seeds ($14$, $42$, and $2027$) for each trained method in \cref{tab:main}, which reports the mean of the run-level scores. The teacher and untouched student are fixed-checkpoint reference evaluations, so their variation is not training-seed variation. Sampling and grading details are in \cref{sec:method-impl}.

\subsection{Main Results}
\label{sec:exp-main}

\begin{figure*}[t]
\centering
\includegraphics[width=0.86\textwidth]{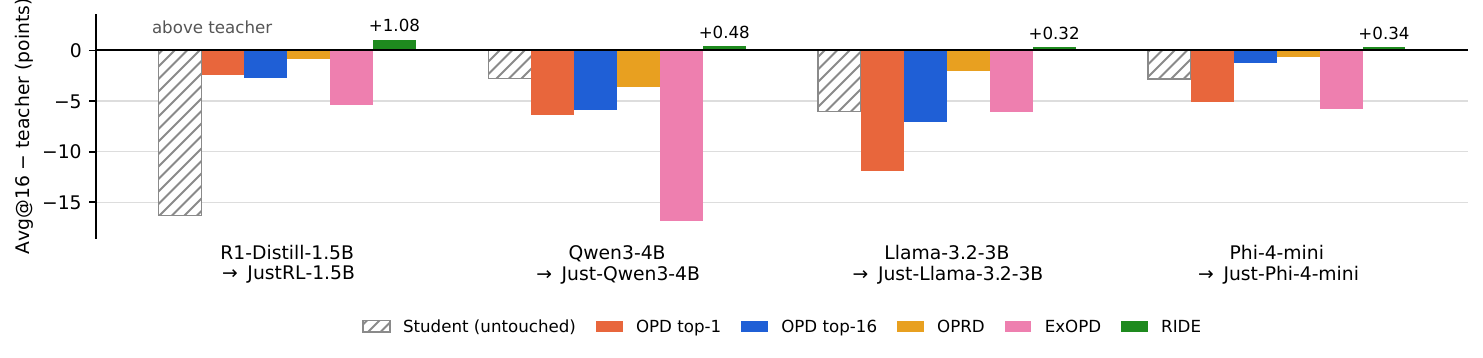}
\caption{Avg@16 relative to the RL-trained teacher (points; the horizontal line is the teacher) for every method and base/teacher pair in \cref{tab:main}. \method{} is the only method whose mean lies above the line on all four pairs; across-seed standard deviations are in \cref{tab:seed-summary}.}
\label{fig:gain}
\end{figure*}

\Cref{tab:main} reports Avg@16 on the four base/RL-teacher pairs, \cref{fig:gain} plots each method's average relative to its teacher, and \cref{tab:seed-summary} reports the across-seed standard deviations for OPRD and \method{}. \method{} is the only method whose mean lies above the teacher line, on every pair, although on three pairs the margin is within one across-seed standard deviation (\cref{tab:seed-summary}); all teacher-matching methods remain below the line, as their objectives dictate, and output-space extrapolation falls below it as well. Since \method{} differs from OPRD only in its target, its margin over OPRD ($0.97$ to $4.06$ points) isolates the benefit of residual extrapolation under a shared protocol.

\Cref{fig:gain} also separates the space in which extrapolation is performed from extrapolation itself: ExOPD and \method{} use the same $\lambda$ and the same pre-RL reference and differ only in where the displacement is measured. ExOPD falls below its teacher on every pair, and on the R1-Distill pair it also falls below the output-space teacher-matching baselines ($49.87$ versus $52.90$ for sampled-token OPD) even though RL moved this teacher $16.3$ points from its base; on the three pairs where that change is small ($2.8$ to $6.0$ points) it falls below the untouched student as well, by $14.1$ points on Qwen3-4B. This is the behavior predicted by \cref{prop:variance}: the extrapolated component of the sampled advantage is scaled by $(\lambda-1)^2$ regardless of how much signal the log-ratio carries, and the damage is largest when the teacher-to-base gap is small. \method{} instead improves over the student and over OPRD on every pair, and over ExOPD by $9.1$ points on average. On the Llama-3.2-3B pair the AIME benchmarks lie near the evaluation floor, so the evidence there rests mainly on AIMO.

\begin{table*}[t]
\centering
\caption{Main results (Avg@16, \%) on four base/RL-teacher pairs. Trained-method entries are means over three independent training seeds; the teacher and untouched student are fixed-checkpoint reference evaluations. Each run uses its own on-policy rollouts; within a pair, all runs share prompts, sampling settings, and optimizer schedule. ExOPD and \method{} use $\lambda=1.25$ with the pre-RL checkpoint as reference. Bold marks the best distillation method per column.}
\label{tab:main}
\small
\setlength{\tabcolsep}{6pt}
\renewcommand{\arraystretch}{0.96}
\begin{tabular}{@{}l*{8}{c}@{}}
\toprule
& \multicolumn{4}{c}{R1-Distill-1.5B $\rightarrow$ JustRL-1.5B}
& \multicolumn{4}{c}{Qwen3-4B $\rightarrow$ Just-Qwen3-4B} \\
\cmidrule(lr){2-5}\cmidrule(l){6-9}
Method & AIME24 & AIME25 & AIMO & Avg.
       & AIME24 & AIME25 & AIMO & Avg. \\
\midrule
Teacher    & 50.80 & 35.60 & 79.50 & 55.30
           & 63.13 & 51.04 & 82.61 & 65.59 \\
Student    & 32.90 & 21.90 & 62.20 & 39.00
           & 61.25 & 48.75 & 78.46 & 62.82 \\
OPD top-1  & 47.10 & 33.50 & 78.10 & 52.90
           & 52.50 & 45.83 & 79.40 & 59.24 \\
OPD top-16 & 47.10 & 34.00 & 76.50 & 52.53
           & 52.92 & 47.08 & 79.22 & 59.74 \\
OPRD       & 49.80 & 34.60 & 79.10 & 54.50
           & 61.46 & 44.38 & 80.20 & 62.01 \\
ExOPD      & 44.20 & 29.60 & 75.80 & 49.87
           & 41.88 & 28.54 & 75.83 & 48.75 \\
\method{}  & \textbf{50.21} & \textbf{37.29} & \textbf{81.63} & \textbf{56.38}
           & \textbf{63.13} & \textbf{52.92} & \textbf{82.15} & \textbf{66.07} \\
\midrule
& \multicolumn{4}{c}{Llama-3.2-3B $\rightarrow$ Just-Llama-3.2-3B}
& \multicolumn{4}{c}{Phi-4-mini $\rightarrow$ Just-Phi-4-mini} \\
\cmidrule(lr){2-5}\cmidrule(l){6-9}
Method & AIME24 & AIME25 & AIMO & Avg.
       & AIME24 & AIME25 & AIMO & Avg. \\
\midrule
Teacher    & 13.75 & 0.42 & 24.85 & 13.01
           & 11.46 & 6.04 & 36.37 & 17.96 \\
Student    & 4.17 & 0.63 & 16.19 & 6.99
           & 7.92 & 4.79 & 32.68 & 15.13 \\
OPD top-1  & 0.63 & 0.21 & 2.41 & 1.08
           & 6.25 & 3.75 & 28.69 & 12.90 \\
OPD top-16 & 2.08 & 0.63 & 15.21 & 5.97
           & 10.00 & 5.63 & 34.64 & 16.76 \\
OPRD       & 7.92 & 1.04 & 23.87 & 10.94
           & \textbf{10.21} & 6.46 & 35.32 & 17.33 \\
ExOPD      & 3.33 & 0.42 & 17.09 & 6.95
           & 7.29 & 1.67 & 27.71 & 12.22 \\
\method{}  & \textbf{13.50} & \textbf{1.30} & \textbf{25.20} & \textbf{13.33}
           & \textbf{10.21} & \textbf{7.50} & \textbf{37.20} & \textbf{18.30} \\
\bottomrule
\end{tabular}
\end{table*}

\subsection{Effect of the Extrapolation Coefficient}
\label{sec:exp-lambda}

\begin{figure*}[t]
\centering
\includegraphics[width=\textwidth]{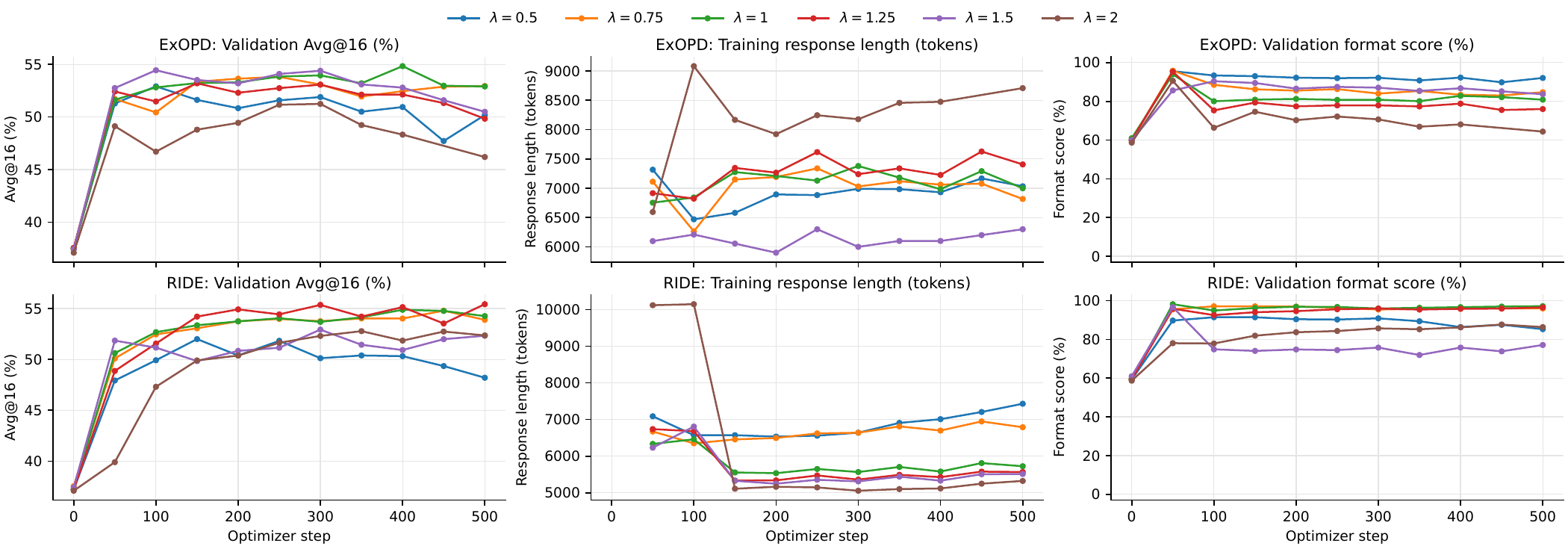}
\caption{ExOPD (top) and \method{} (bottom) at six extrapolation coefficients on the R1-Distill-1.5B pair (one run each): validation Avg@16 (\%) averaged over AIME24, AIME25, and AIMO; mean response length over the preceding $50$ steps; validation format score.}
\label{fig:lambda}
\end{figure*}

\Cref{fig:lambda} compares \method{} and ExOPD on the R1-Distill-1.5B pair for $\lambda\in\{0.5,0.75,1,1.25,1.5,2\}$; at $\lambda=1$ they reduce to OPRD and to sampled-token OPD, so each row isolates extrapolation within one space. For \method{}, final Avg@16 rises with $\lambda$ from $48.2$ at $\lambda=0.5$ to $54.3$ at $1$ and peaks at $55.4$ at $1.25$; the neighboring coefficients $\lambda=1.15$ and $1.35$ give $55.2$ and $55.3$, so every $\lambda\in[1.15,1.35]$ improves over OPRD (full numbers in \cref{app:lambda}). Every run with $0.75\le\lambda\le1.35$ improves steadily over the whole budget with a format score of $94$--$97\%$, and beyond $\lambda=1.35$ the degradation is graceful, with $\lambda=1.5$ and $2$ finishing at $52.3$ and $52.4$. Displacing beyond the teacher ($\lambda\ge1$) shortens responses to $5.3$--$5.7$k tokens, whereas displacing toward the base ($\lambda<1$) lengthens them to $6.8$--$7.4$k. ExOPD is best at $\lambda\le1$ ($52.9$ at $0.75$ and at $1$) and is harmed by every $\lambda>1$: the extrapolated runs reach their peak early (by step $150$ for $\lambda=1.25$ and $1.5$) and then decline ($\lambda=1.25$ to $49.9$, $\lambda=2$ to $46.2$), the format score falls from $92\%$ at $\lambda=0.5$ to $64\%$ at $2$, and responses lengthen to $8{,}700$ tokens. This is the behavior predicted by \cref{prop:variance}: the sampled-advantage noise scales with $(\lambda-1)^2$, so the coefficient that moves \method{} further along the RL-induced direction moves ExOPD off the teacher's formatting and below its teacher-matching baseline. Elsewhere we use $\lambda=1.25$ for both methods, the best value for \method{} and the smallest extrapolating value for ExOPD.

\subsection{Mechanistic Analysis}
\label{sec:exp-mechanism}

\begin{figure*}[t]
\centering
\begin{subfigure}[t]{0.33\linewidth}
  \centering
  \includegraphics[width=\linewidth]{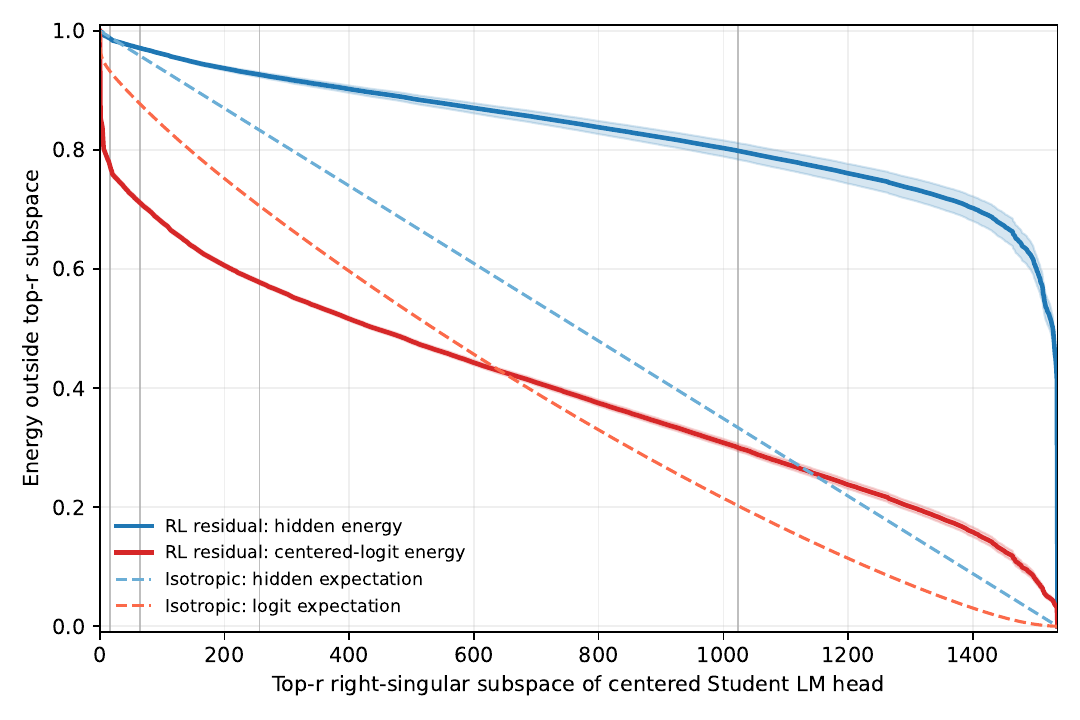}
  \caption{Residual energy outside top-$r$ head directions.}
\end{subfigure}\hfill
\begin{subfigure}[t]{0.24\linewidth}
  \centering
  \includegraphics[width=\linewidth]{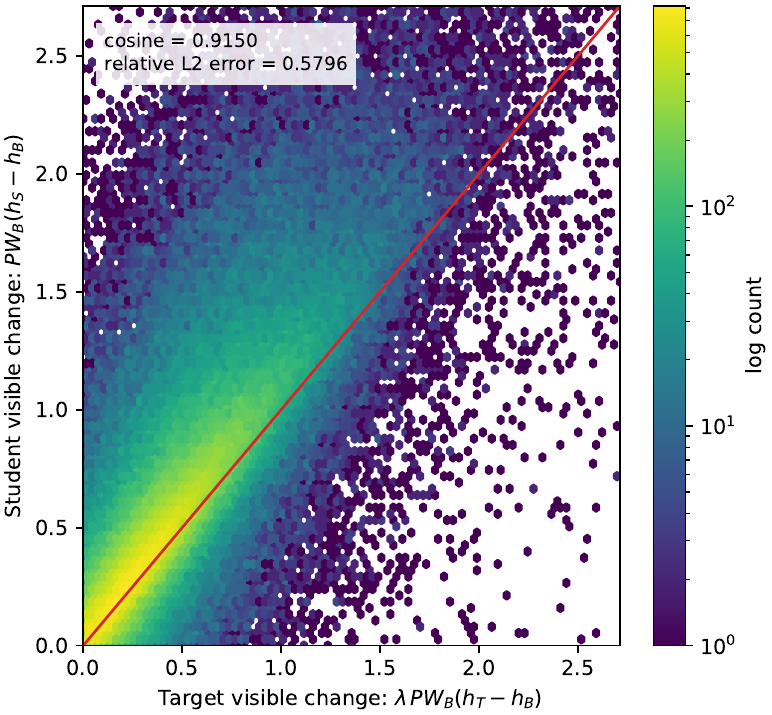}
  \caption{Student versus target change.}
\end{subfigure}\hfill
\begin{subfigure}[t]{0.33\linewidth}
  \centering
  \includegraphics[width=\linewidth]{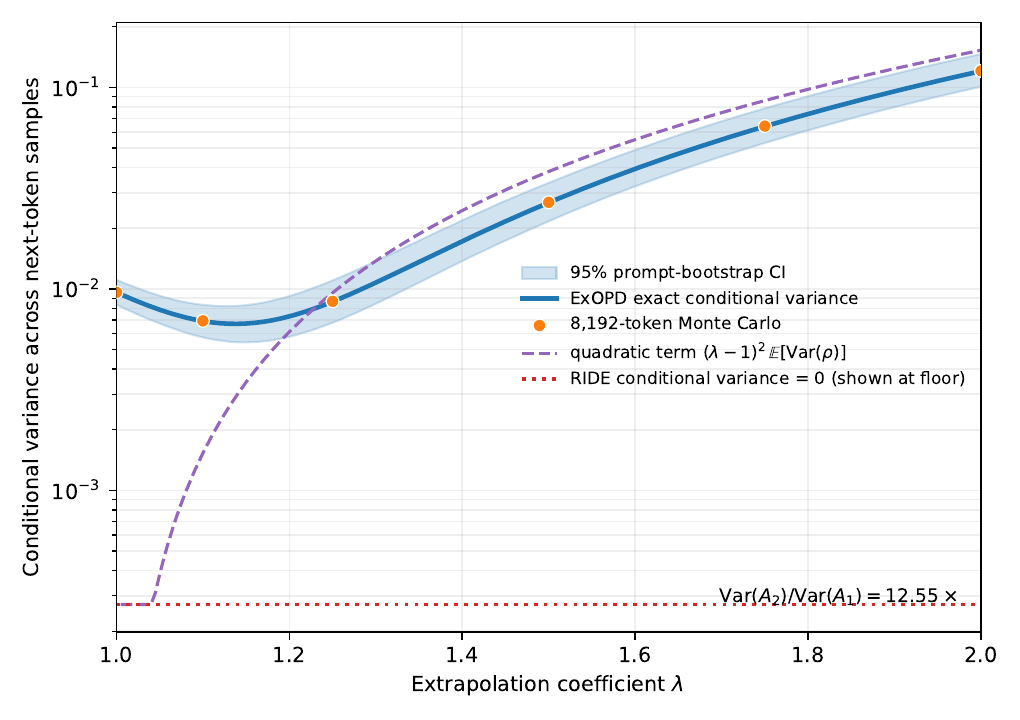}
  \caption{Conditional variance across $\lambda$.}
\end{subfigure}
\caption{Mechanistic analysis on the R1-Distill-1.5B pair ($\lambda=1.25$, step $500$, final layer). (a) The $512$ weakest head directions hold $79.8\%$ of the residual's hidden-state energy versus $33.3\%$ under isotropy. (b) Realized student change versus target. (c) Next-token conditional variance of the ExOPD advantage and the \method{} gradient norm across $\lambda$.}
\label{fig:head}
\end{figure*}

\paragraph{Head attenuation reaches the learning signal.}
The residual retains only $0.59$ of the head gain of an equal-norm isotropic direction, and its energy concentrates in the weakest head directions (\cref{fig:head}a); these directions receive $73.0\%$ of the representation-loss gradient but only $48.6\%$ of the output-KL gradient at the head input, so head attenuation weakens the output-space learning signal as well. The centered head is full rank, with singular values spanning $71.7$ to $1.8$, so the residual is attenuated by a factor of up to $40$ along its dominant directions rather than removed: an output-space objective still sees the RL-induced change, but at a fraction of its weight. The student's update aligns with the residual (cosine $0.954$) and its head-visible component tracks the target (\cref{fig:head}b, cosine $0.895$), with a projection onto the residual of $1.69$, above the target coefficient of $1.25$: $\lambda$ specifies the target rather than the realized displacement, and the student continues past the target along the same direction. During RL the teacher head itself drifts from $W_B$ by only $1.65\%$, which supports the shared-head assumption behind \cref{eq:logit-image}; the representation-space target contains no head-drift term. Further measurements are in \cref{app:head-diagnostics}.

\paragraph{The learning signal is deterministic in representation space.}
At a fixed prefix, the sampled-token advantage of ExOPD varies with the next token, whereas the \method{} gradient does not (\cref{prop:variance}). On $9{,}152$ on-policy validation prefixes per method (\cref{fig:head}c), the conditional variance of ExOPD rises $12.6\times$ from $\lambda=1$ to $\lambda=2$, with the extrapolation term dominating beyond $\lambda\approx1.3$, whereas resampling the next token leaves the \method{} per-position gradient norm unchanged at every prefix and every $\lambda$ (drawn at the floor of the logarithmic axis). The decomposition of \cref{eq:var} also explains why ExOPD tolerates coefficients slightly above one: the covariance term is negative, so the variance has a shallow minimum near $\lambda\approx1.14$ before the $(\lambda-1)^2\,\mathrm{Var}[\rho]$ term takes over, which matches the sweep in \cref{sec:exp-lambda}, where ExOPD degrades at $\lambda=1.25$ and collapses at $\lambda=2$. The sampling protocol, variance decomposition, and uncertainty estimates are in \cref{app:signal-diagnostics}.

\subsection{Ablation: The Residual Direction}
\label{sec:exp-ablation}

A displaced target could help for reasons unrelated to the RL-induced residual, for instance by acting as a regularizer or by enlarging the loss scale. We therefore ablate the residual direction on the R1-Distill-1.5B pair with four controls that hold the displacement magnitude fixed and alter only its direction (\cref{tab:direction} in \cref{app:direction}). The \emph{random} control replaces $\Delta^{(l)}_t$ with a Gaussian vector of the same norm; the \emph{reversed} control uses $\lambda<1$, displacing the target toward $\pi_B$; the \emph{mismatched-origin} control computes the residual against Qwen2.5-Math-1.5B-Instruct \citep{yang2024qwen25math}, which shares the architecture but not the training lineage, so that $h_T-h_{B'}$ reflects a difference between unrelated models rather than the change produced by RL; and the \emph{trajectory-mismatched} control computes $h_T$ and $h_B$ on separately generated trajectories rather than on the identical student prefix, testing the claim in \cref{sec:method-residual} that the fixed context gives the residual its meaning. Relative to OPRD ($54.50$), the random, reversed, mismatched-origin, and trajectory-mismatched controls reach $55.04$, $53.90$, $54.75$, and $55.12$, all within one point, whereas \method{} with the RL-induced residual reaches $56.38$.

\section{Conclusion}
\label{sec:conclusion}

An RL-trained teacher supplies a direction as well as a destination. The layerwise hidden-state difference between the teacher and its pre-RL checkpoint on identical prefixes measures what RL changed, and \method{} displaces the representation-matching target beyond the teacher along it; only the target changes, so $\lambda=1$ recovers OPRD. Measuring it before the head matters because the head attenuates the residual anisotropically and leaves earlier layers unconstrained, and because output-space extrapolation amplifies sampled-token noise by $(\lambda-1)^2$ whereas the \method{} gradient is deterministic given the rollout. Across four base/teacher pairs, \method{} approaches or exceeds its teacher, consistently outperforms OPRD and output-space extrapolation, remains stable where the latter collapses, and, as the direction controls show, owes its gain to the RL-induced residual itself. \method{} requires the pre-RL checkpoint and a shared representation space, uses a single global $\lambda$, and was evaluated only on mathematical reasoning with one RL recipe. Relaxing these constraints and studying the safety and calibration of students trained on hidden-state targets are left for future work.

\FloatBarrier
\subsection*{Ethics statement}
This work studies a training objective for distilling mathematical-reasoning ability from an RL-trained teacher into a student that shares its initialization. All experiments use publicly released model checkpoints and public prompt and evaluation sets (DAPO-Math-17K, AIME 2024, AIME 2025, and the AIMO AMC 2022--2023 problems); no human subjects, personal data, or proprietary data are involved, and the task domain (competition mathematics) carries no direct risk of harm. Two considerations nonetheless apply. First, like any distillation method, \method{} transfers the teacher's behavior, and therefore also any biases or errors that RL introduced into it, to the student; because \method{} displaces the target \emph{beyond} the teacher, such properties could in principle be amplified rather than merely copied. Second, the student is supervised on hidden-state targets that no model has actually produced, so its calibration and safety behavior are not guaranteed to match those of the teacher and should be evaluated independently before deployment; we note this limitation in \cref{sec:conclusion}. We do not foresee misuse specific to this method beyond that of on-policy distillation in general, and we report the compute used (\cref{tab:hparams}) so that the environmental cost of reproducing our results can be assessed.

\subsection*{Reproducibility statement}
We have taken the following steps to make our results reproducible. The \method{} objective is fully specified by \cref{eq:target,eq:ride} together with the loss rescaling in \cref{rem:scale}, and its implementation on top of an OPRD pipeline is described in \cref{sec:method-impl}, including how the pre-RL checkpoint is hosted and how the target is formed and transmitted. \Cref{tab:hparams} lists every training and evaluation hyperparameter shared by all methods (prompt set, rollout settings, optimizer, schedule, precision, hardware, supervised layers and positions, $\lambda$, evaluation temperature, and benchmark sizes), and \cref{tab:cost} states the per-rollout cost of each method. All base checkpoints and the JustRL-DeepSeek-1.5B teacher are publicly available; the remaining teachers are obtained by applying the published JustRL recipe \citep{he2025justrl} to the corresponding public base models, as described in \cref{sec:exp-setup}. Complete proofs of \cref{prop:variance,prop:reward} and the derivation of the extrapolated advantage are given in \cref{app:proofs}. The full extrapolation-coefficient sweep behind \cref{fig:lambda}, including per-benchmark numbers, peak checkpoints, format scores, and response lengths, is tabulated in \cref{tab:lambda}, and the sampling protocols and sample sizes for every diagnostic in \cref{sec:exp-mechanism} are specified in \cref{app:head-diagnostics,app:signal-diagnostics}. Training code, the RL-trained teacher checkpoints, and evaluation scripts will be released upon publication.

\subsection*{AI use statement}
In this work, we used generative AI tools to assist with language polishing, checking consistency of terminology and notation, reviewing the presentation and interpretation of reported results, and verifying bibliographic metadata against source records. The tools did not design or run experiments or generate the reported measurements, figures, or tables. The authors are responsible for the research decisions, technical claims, and final manuscript, including AI-assisted edits.

\bibliography{iclr2027_conference}
\bibliographystyle{iclr2027_conference}

\appendix
\section{Derivations and Proofs}
\label{app:proofs}

\subsection{On-Policy Distillation as Implicit-Reward Reinforcement Learning}
\label{app:opd}

Fix a prefix and write $p=\pi_\theta(\cdot\mid x,\hat y_{<t})$, $q=\pi_T(\cdot\mid x,\hat y_{<t})$, and let $\pi_{\mathrm{ref}}$ be any reference policy with full support. Adding and subtracting $\log\pi_{\mathrm{ref}}$ inside the local KL of \cref{eq:opd} gives
\begin{equation}
D_{\mathrm{KL}}(p\,\|\,q)=D_{\mathrm{KL}}(p\,\|\,\pi_{\mathrm{ref}})-\mathbb{E}_{v\sim p}\big[\log q(v)-\log\pi_{\mathrm{ref}}(v)\big],
\end{equation}
so minimizing the OPD loss is KL-regularized RL with the implicit reward $\log q-\log\pi_{\mathrm{ref}}$, in which the reward and the KL term have equal weight \citep{yang2026learning}. Weighting the reward by $\lambda$ instead, the local objective $\lambda\,\mathbb{E}_{v\sim p}[\log q-\log\pi_{\mathrm{ref}}]-D_{\mathrm{KL}}(p\,\|\,\pi_{\mathrm{ref}})$ is maximized by
\begin{equation}
\log p^\star=\lambda\log q+(1-\lambda)\log\pi_{\mathrm{ref}}-\log Z,
\end{equation}
where $Z$ normalizes $p^\star$. With $\pi_{\mathrm{ref}}=\pi_B$ this is \cref{eq:principle} on log-probabilities. Minimizing $D_{\mathrm{KL}}(p\,\|\,p^\star)$ with the sampled-token estimator of \cref{eq:opd-grad} uses the per-token quantity
\begin{equation}
\log p(v)-\log p^\star(v)=\ell(v)-(\lambda-1)\rho(v)+\log Z,
\end{equation}
with $\ell$ and $\rho$ as in \cref{sec:method-why}. The constant $\log Z$ does not depend on $v$ and multiplies a score function with zero conditional mean, so the extrapolated advantage is $A_\lambda(v)=\ell(v)-(\lambda-1)\rho(v)$. The tokenwise update treats sampled prefixes as fixed; differentiating the sequence-level objective in \cref{eq:opd} would additionally account for how earlier actions change later prefixes.

\subsection{Proof of \texorpdfstring{\cref{prop:variance}}{Proposition}}

For random variables $X,Y$ and a constant $c$, $\mathrm{Var}[X-cY]=\mathrm{Var}[X]+c^2\mathrm{Var}[Y]-2c\,\mathrm{Cov}[X,Y]$. Taking $X=\ell(v)$, $Y=\rho(v)$, $c=\lambda-1$, and $v\sim p$ gives \cref{eq:var}. Since $q$ and $b$ are frozen, $\rho$ does not depend on $\theta$, and the second term grows quadratically in $\lambda-1$. As $p\to q$, $\mathrm{Var}_{v\sim p}[\rho(v)]\to\mathrm{Var}_{v\sim q}[\rho(v)]$, which vanishes only if $\log q-\log b$ is constant on the support of $q$, that is, only if RL left the next-token distribution at this prefix unchanged.

For \method{}, the student state $h^{(l)}_{\theta,t}$ is a deterministic function of $\theta$ and the prefix, and the target $h^{\star(l)}_t$ is a deterministic function of the prefix and the two frozen models. Neither depends on the token sampled at position $t$, so the per-position loss and its gradient with respect to $\theta$ are invariant to resampling that token, and their conditional variance is zero.

\subsection{Reward-and-Penalty Form: Derivation of \texorpdfstring{\cref{prop:reward}}{Proposition}}
\label{app:reward}

With $h$, $h_T$, and $\Delta$ as in \cref{sec:method-analysis}, expanding the squared distance to the target gives
\begin{equation}
\label{eq:expand}
\big\|h-h_T-(\lambda-1)\Delta\big\|_2^2
=\big\|h-h_T\big\|_2^2-2(\lambda-1)\big\langle h-h_T,\;\Delta\big\rangle+(\lambda-1)^2\|\Delta\|_2^2 ,
\end{equation}
in which the final term is independent of $\theta$. The squared distance therefore equals $\|h-h_T\|_2^2-2(\lambda-1)\,r(h)$ plus a constant, and multiplying by $-\tfrac12$ gives \cref{eq:reward-form} up to an additive constant; the per-position loss is this squared distance scaled by the positive factor $1/d$. The objective in \cref{eq:reward-form} is strictly concave in $h$ with gradient $(\lambda-1)\Delta-(h-h_T)$, which vanishes at $h^\star=h_T+(\lambda-1)\Delta$. Differentiating $\tfrac1d\|h-h^\star\|_2^2$ gives $\tfrac{2}{d}(h-h^\star)$.

The reward $r$ measures the extent to which the student's displacement from the teacher is aligned with the RL-induced residual, the penalty is quadratic in the same displacement, and $\lambda-1$ weights the reward against the penalty. This mirrors the view of output-space extrapolation as KL-constrained RL around the teacher \citep{yang2026learning}, with the KL constraint replaced by a quadratic one, the log-ratio reward by an inner product with the residual, and the score-function estimator by a vector gradient. At $\lambda=1$ the reward weight vanishes: teacher matching is a special case of \method{} rather than its point of departure.

\section{Extended Extrapolation-Coefficient Sweep}
\label{app:lambda}

\begin{figure}[h]
\centering
\includegraphics[width=\textwidth]{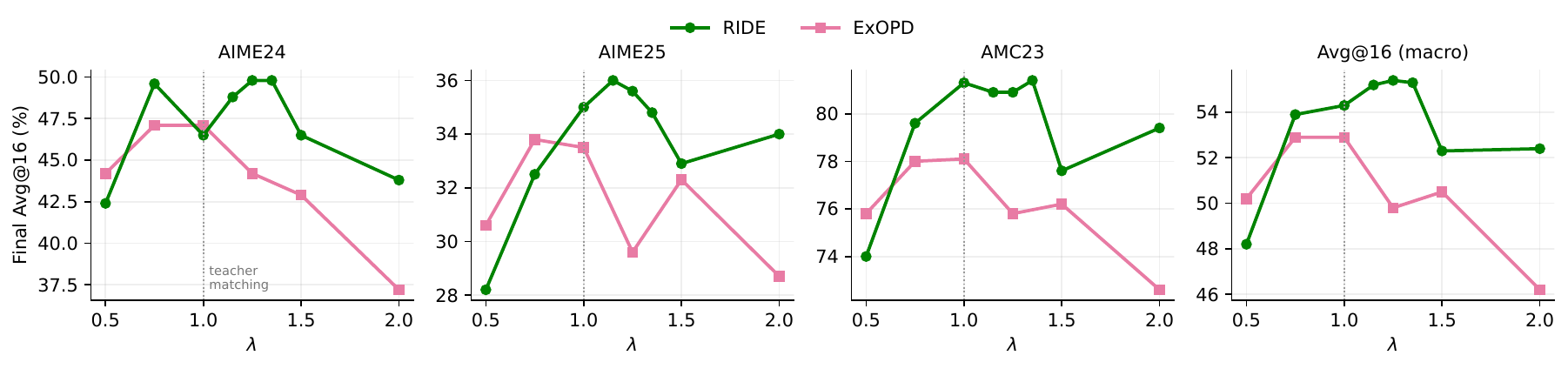}
\caption{Final Avg@16 (\%) per benchmark as a function of the extrapolation coefficient on the R1-Distill-1.5B pair (one run per coefficient). The dotted line marks $\lambda=1$, which is OPRD for \method{} and sampled-token OPD for ExOPD.}
\label{fig:appx-lambda-bench}
\end{figure}

\begin{figure}[h]
\centering
\includegraphics[width=0.95\textwidth]{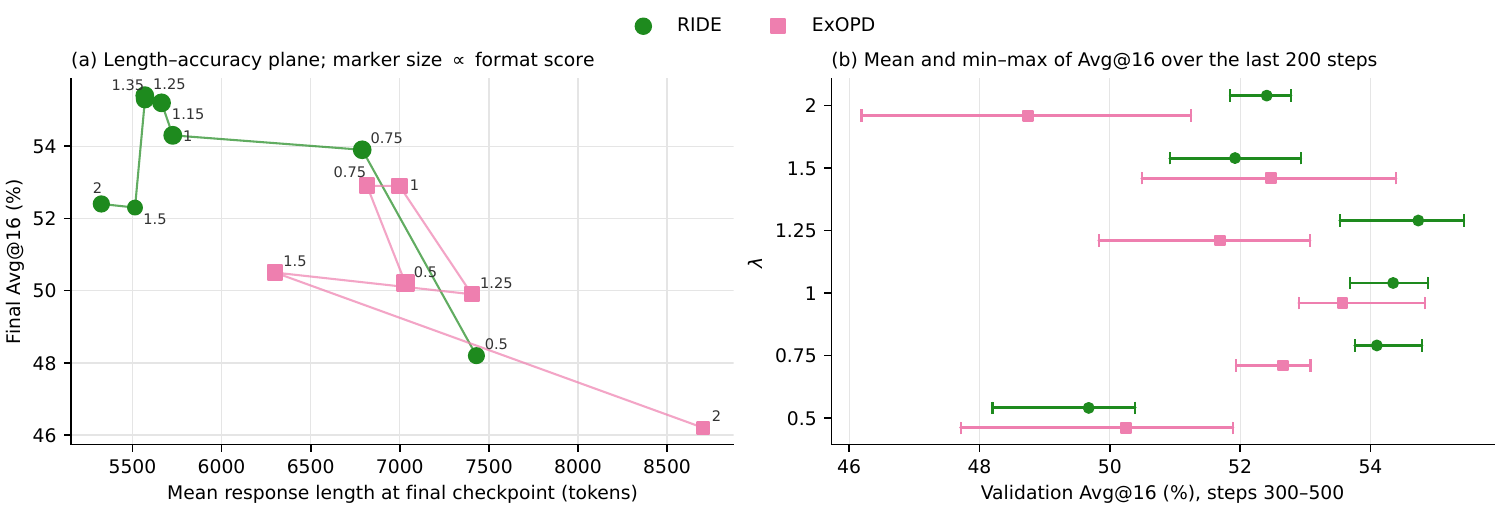}
\caption{Extrapolation-coefficient sweep viewed per run. (a) Final Avg@16 against mean response length at the final checkpoint; markers are labeled by $\lambda$, connected in increasing $\lambda$, and sized by validation format score. (b) Mean (marker) and min--max range (bar) of validation Avg@16 over the last $200$ training steps (steps $300$--$500$) for the six coefficients tracked in \cref{fig:lambda}.}
\label{fig:appx-lambda-plane}
\end{figure}

\Cref{tab:lambda} lists the numbers behind \cref{fig:lambda}: validation Avg@16 on AIME24, AIME25, and AIMO at the final checkpoint, the validation format score, and the mean training response length over the $50$ steps preceding the final checkpoint. The two additional \method{} coefficients $\lambda=1.15$ and $1.35$ are included; they bracket $\lambda=1.25$ and both remain above OPRD ($55.2$ and $55.3$ versus $54.3$), so every coefficient in $[1.15,1.35]$ improves over teacher matching, with $\lambda=1.25$ the best. The sweep uses a single training seed per coefficient, whereas \cref{tab:main} averages three seeds; the $\lambda=1$ and $\lambda=1.25$ rows of \cref{tab:lambda} are therefore single runs ($54.3$ and $55.4$) and differ from the corresponding three-seed means in \cref{tab:main} ($54.50$ and $56.38$) by about one across-seed standard deviation (\cref{tab:seed-summary}). \Cref{fig:appx-lambda-bench} shows the final Avg@16 on each benchmark as a function of $\lambda$: the gain of \method{} around $\lambda=1.25$ and the decline of ExOPD beyond $\lambda=1$ appear on all three benchmarks, not only in the macro average. \Cref{fig:appx-lambda-plane}a places every run in the length--accuracy plane; increasing $\lambda$ moves \method{} toward shorter and more accurate responses with a high format score, whereas it moves ExOPD toward longer and less accurate ones. \Cref{fig:appx-lambda-plane}b summarizes the last $200$ training steps of each run in \cref{fig:lambda} by the mean and the min--max range of validation Avg@16 over steps $300$--$500$. For \method{}, the late-training mean rises from $54.3$ at $\lambda=1$ to $54.7$ at $\lambda=1.25$, and every run with $\lambda\ge0.75$ stays above $50.9$ throughout the window; for ExOPD, the late-training mean of every $\lambda>1$ run lies below that of $\lambda=1$ ($53.6$), the ranges widen to $3.2$--$5.1$ points, and the $\lambda=2$ run drifts from $51.2$ to $46.2$ within the window, so the coefficient that improves \method{} degrades ExOPD and makes its late-training behavior less stable. For \method{}, every $\lambda\ge1.15$ shortens responses relative to OPRD while $\lambda\le0.75$ lengthens them; for ExOPD, the final Avg@16 is non-monotone in $\lambda$ and every $\lambda>1$ ends below the $\lambda=1$ run, and the format score decreases with $\lambda$ except for a partial recovery at $\lambda=1.5$.

\begin{table}[h]
\centering
\caption{Extrapolation-coefficient sweep on the R1-Distill-1.5B pair (one run per coefficient). Per-benchmark and Avg@16 columns are validation Avg@16 (\%) at the final checkpoint; ``Format'' is the validation format score (\%); ``Length'' is the mean response length (tokens) over the $50$ steps preceding the final checkpoint. Bold marks the best final Avg@16 within each method.}
\label{tab:lambda}
\small
\setlength{\tabcolsep}{3.4pt}
\begin{tabular}{l c ccc c c c}
\toprule
Method & $\lambda$ & AIME24 & AIME25 & AIMO & Avg@16 & Format & Length \\
\midrule
\method{} & 0.5 & 42.4 & 28.2 & 74.0 & 48.2 & 85.3 & 7430 \\
 & 0.75 & 49.6 & 32.5 & 79.6 & 53.9 & 96.0 & 6789 \\
 & 1 (OPRD) & 46.5 & 35.0 & 81.3 & 54.3 & 97.2 & 5726 \\
 & 1.15 & 48.8 & 36.0 & 80.9 & 55.2 & 96.4 & 5663 \\
 & 1.25 & 49.8 & 35.6 & 80.9 & \textbf{55.4} & 96.6 & 5569 \\
 & 1.35 & 49.8 & 34.8 & 81.4 & 55.3 & 94.2 & 5570 \\
 & 1.5 & 46.5 & 32.9 & 77.6 & 52.3 & 77.1 & 5514 \\
 & 2 & 43.8 & 34.0 & 79.4 & 52.4 & 86.4 & 5325 \\
\midrule
ExOPD & 0.5 & 44.2 & 30.6 & 75.8 & 50.2 & 92.1 & 7032 \\
 & 0.75 & 47.1 & 33.8 & 78.0 & \textbf{52.9} & 84.7 & 6816 \\
 & 1 (OPD) & 47.1 & 33.5 & 78.1 & \textbf{52.9} & 80.9 & 6998 \\
 & 1.25 & 44.2 & 29.6 & 75.8 & 49.9 & 76.0 & 7404 \\
 & 1.5 & 42.9 & 32.3 & 76.2 & 50.5 & 83.7 & 6300 \\
 & 2& 37.2 & 28.7 & 72.6 & 46.2 & 64.4 & 8704 \\
\bottomrule
\end{tabular}
\end{table}

\FloatBarrier

\section{Direction Controls}
\label{app:direction}

\Cref{tab:direction} lists the direction-control results discussed in \cref{sec:exp-ablation}. Every control keeps the training protocol, the displacement magnitude $|\lambda-1|\,\|\Delta^{(l)}_t\|$, and the layer and position sets of \method{} fixed and changes only the direction of the displacement.

\begin{table}[h]
\centering
\caption{Direction controls at fixed displacement magnitude; Avg@16 (\%) averaged over AIME24, AIME25, and AIMO on the R1-Distill-1.5B pair; $\lambda=1.25$ unless stated.}
\label{tab:direction}
\small
\begin{tabular}{l c}
\toprule
Target construction & Avg@16 \\
\midrule
Teacher matching (OPRD, $\lambda=1$) & 54.50 \\
\midrule
Random direction, same norm & 55.04 \\
Reversed direction ($\lambda=0.75$) & 53.90 \\
Mismatched origin ($h_T-h_{B'}$) & 54.75 \\
Trajectory-mismatched residual & 55.12 \\
\midrule
\method{} (RL-induced residual) & 56.38 \\
\bottomrule
\end{tabular}
\end{table}

\section{Training-Seed Summary}
\label{app:seeds}

\Cref{tab:seed-summary} summarizes the main-table macro Avg@16 means and across-seed standard deviations for OPRD and \method{} on the four base/RL-teacher pairs. Each run-level macro score averages the three benchmark Avg@16 scores with equal weight, as in \cref{sec:exp-setup}. The three independent training seeds are $14$, $42$, and $2027$.

\begin{table}[h]
\centering
\caption{Mean $\pm$ SD of macro Avg@16 (\%) over three independent training runs (seeds $14$, $42$, and $2027$). Means are taken from \cref{tab:main}; SDs are across training seeds. Teacher scores are fixed-checkpoint references.}
\label{tab:seed-summary}
\small
\setlength{\tabcolsep}{5pt}
\begin{tabular}{l c c c}
\toprule
Base/RL-teacher pair & Teacher & OPRD & \method{} \\
\midrule
R1-Distill-1.5B & 55.30 & $54.50 \pm 1.04$ & $56.38 \pm 0.97$ \\
Qwen3-4B & 65.59 & $62.01 \pm 0.96$ & $66.07 \pm 0.88$ \\
Llama-3.2-3B & 13.01 & $10.94 \pm 0.90$ & $13.33 \pm 1.13$ \\
Phi-4-mini & 17.96 & $17.33 \pm 1.06$ & $18.30 \pm 1.05$ \\
\bottomrule
\end{tabular}
\end{table}

\section{Additional Head Diagnostics}
\label{app:head-diagnostics}

The measurements in \cref{fig:head} use $292{,}864$ positions from the R1-Distill-1.5B pair: $143$ problems, $16$ rollouts each, and $128$ positions per rollout from the last $2{,}000$ response tokens at step $500$. The residual's normalized head gain is $0.59$ (prompt-bootstrap $95\%$ CI $[0.585,0.596]$), and its hidden-state energy outside the top-$r$ head subspace exceeds the isotropic expectation at every $r$. The centered head is full rank, with singular values ranging from $71.70$ to $1.78$, so the residual is attenuated rather than removed. Across training steps, the $512$ weakest directions carry $73.00\%$ of the representation-loss gradient of \cref{eq:ride}, versus $48.60\%$ of the output-KL gradient at the head input.

\paragraph{Scope of \cref{eq:logit-image}.} \Cref{eq:ride} supervises the pre-normalization residual stream, so \cref{eq:logit-image} holds after the final normalization is applied to the target; the identity is stated at the head input so that it holds exactly. It also assumes the head shared by base and student: a teacher whose head drifted during RL contributes an additional term $(W_T-W_B)\tilde h_T$ to its native output-space target, quantified below. With $W_B$ fixed, the head image of the \method{} target matches $\lambda z_T+(1-\lambda)z_B$ to relative error $6.50\times10^{-7}$. During RL, the teacher head drifts from $W_B$ by $1.65\%$ in relative Frobenius norm. The resulting term $(W_T-W_B)\tilde h_T$ has $50\%$ of the norm of the native logit change, compared with $94\%$ for the representation term, and the two have cosine similarity $-0.23$. Fixed-head and native-head targets differ by $18.70\%$ in mean relative logit error, although the corresponding distributions have a mean KL divergence of $0.0058$ and a top-$1$ agreement of $96.90\%$. The representation-space target contains no head-drift term. For the student update in \cref{fig:head}(b), the residual-alignment cosine is $0.954$ (prompt-bootstrap $95\%$ CI $[0.949,0.958]$), the head-visible cosine is $0.895$ over $5.20\times10^{5}$ coordinates, and the mean projection along the residual is $1.69$ (CI $[1.65,1.73]$).

\section{Additional Learning-Signal Diagnostics}
\label{app:signal-diagnostics}

From each method's on-policy validation rollouts, we use $143$ problems, $4$ rollouts per problem, and $16$ positions per rollout drawn uniformly from the last $2{,}000$ response tokens, giving $9{,}152$ fixed prefixes per method. For ExOPD, summing over the full vocabulary of $151{,}936$ tokens at each prefix gives the exact conditional moments $\mathrm{Var}[\ell]$, $\mathrm{Var}[\rho]$, and $\mathrm{Cov}[\ell,\rho]$, and hence $\mathrm{Var}_{v\sim p}[A_\lambda]$ for any $\lambda$. Resampling the next token $8{,}192$ times per prefix reproduces the exact values to within $0.38\%$. For \method{}, we measure $\|\nabla_h\mathcal{L}_\lambda\|_2$ at each fixed prefix and resample the next token $8{,}192$ times per prefix ($7.50\times10^{7}$ draws in total, with adjacent draws differing in $19.50\%$ of cases).

Averaged over prefixes, the ExOPD conditional variance follows \cref{eq:var} with fitted coefficients
\begin{equation*}
\mathrm{Var}[A_\lambda]\;=\;0.0096\;-\;0.0421\,(\lambda-1)\;+\;0.153\,(\lambda-1)^2 .
\end{equation*}
The terms are $\mathrm{Var}[\ell]=0.0096$, $-2\,\mathrm{Cov}[\ell,\rho]=-0.0421$, and $\mathrm{Var}[\rho]=0.153$. The negative covariance term produces a minimum of $0.0067$ at $\lambda\approx1.14$; beyond this point the quadratic term dominates, giving $2.80\times$ the $\lambda=1$ variance at $\lambda=1.50$ and $12.60\times$ at $\lambda=2$ ($0.121$ versus $0.0096$; prompt-bootstrap $95\%$ CIs $[0.101,0.146]$ and $[0.0084,0.0111]$, respectively). In \cref{fig:head}c, shading shows these confidence intervals and the dashed curve isolates $(\lambda-1)^2\,\mathrm{Var}[\rho]$, which accounts for essentially all growth beyond $\lambda\approx1.30$.

For \method{}, the target $h^\star$ depends on the prefix and the frozen models but not on the token drawn next. The variance of its per-position gradient norm across resampled tokens is therefore exactly zero at every prefix and every $\lambda$; the red curve is placed at the plotting floor and does not represent a measured positive value. The gradient norm does vary across prefixes (between-prefix variance $\approx4\times10^{3}$ at every $\lambda$), which reflects variation in the data rather than next-token sampling noise.

\section{Implementation and Diagnostics}
\label{sec:method-impl}

\paragraph{Training and evaluation configuration.}
\Cref{tab:hparams} lists the configuration shared by all methods; only the target construction differs between runs. Final answers are extracted with the standard boxed parser and graded by exact match. \Cref{tab:cost} summarizes the per-rollout cost of each method relative to OPRD: \method{} adds one frozen forward pass of $\pi_B$ on the teacher worker and no additional communication, since $h^\star$ is formed in place and a single hidden-state tensor is transmitted, as for OPRD.

\begin{table}[h]
\centering
\caption{Training and evaluation configuration shared by all methods and all four base/teacher pairs.}
\label{tab:hparams}
\small
\begin{tabular}{l l}
\toprule
\multicolumn{2}{l}{\textit{Rollout}} \\
Prompt dataset & DAPO-Math-17K \\
Prompts per step / responses per prompt & $8$ / $2$ \\
Sampling temperature & $1.00$ \\
Maximum response length & $16{,}384$ tokens \\
\midrule
\multicolumn{2}{l}{\textit{Optimization}} \\
Steps & $500$ \\
Optimizer / peak learning rate & AdamW / $1\times10^{-5}$ \\
Schedule & $3\%$ linear warm-up, cosine decay \\
Precision / parallelism & bf16 / FSDP (verl) \\
Hardware & $4\times$ H200 \\
\midrule
\multicolumn{2}{l}{\textit{\method{} target}} \\
Supervised layers / positions & all $L$ layers / last $k=2{,}000$ positions \\
Extrapolation coefficient $\lambda$ & $1.25$ (sweep in \cref{app:lambda}) \\
Loss coefficient & scaled by $\lambda^{-2}$ (\cref{rem:scale}) \\
Reference checkpoint & pre-RL base $\pi_B$ (frozen) \\
\midrule
\multicolumn{2}{l}{\textit{Evaluation}} \\
Metric & Avg@16, temperature $0.70$ \\
Benchmarks (problems) & AIME24 ($30$), AIME25 ($30$), AIMO ($83$; AMC 2022--2023) \\
\bottomrule
\end{tabular}
\end{table}

\begin{table}[h]
\centering
\caption{Per-rollout cost of each method. ``Frozen passes'' counts forward passes of frozen models on the student's rollout; ``Reference'' is whether the pre-RL checkpoint must be loaded; ``To trainer'' is the supervision tensor transmitted from the teacher worker per rollout ($V$: vocabulary size; $L$, $d$: supervised layers and hidden width).}
\label{tab:cost}
\small
\begin{tabular}{l c c l l}
\toprule
Method & Frozen passes & Reference & To trainer & Target computed \\
\midrule
OPD top-1 / top-16 & $1$ (teacher) & no & $T\times1$ / $T\times16$ log-probs & on trainer \\
OPRD & $1$ (teacher) & no & $T\times L\times d$ hidden states & on teacher worker \\
ExOPD & $2$ (teacher, base) & yes & $T\times1$ log-ratios & on trainer \\
\method{} & $2$ (teacher, base) & yes & $T\times L\times d$ hidden states & on teacher worker \\
\bottomrule
\end{tabular}
\end{table}

\paragraph{Target construction.}
The frozen pre-RL checkpoint is loaded into the worker that already hosts the teacher. For each rollout, this worker runs one additional forward pass of $\pi_B$, forms $h^\star$ in place according to \cref{eq:target}, and transmits a single hidden-state tensor to the trainer, so the transfer volume equals that of OPRD. The loss, layer alignment, and position mask are those of OPRD; positions outside the mask or beyond the response length are zero-padded in both hidden-state tensors, so $h^\star$ vanishes there and the masked loss is unaffected. When $\lambda=1$ or no base checkpoint is provided, the base model is neither loaded nor evaluated, and training coincides with OPRD.

\begin{remark}[Loss scale]
\label[remark]{rem:scale}
At initialization the student coincides with the base, $h_{\theta_0}=h_B$, so $h_{\theta_0}-h^\star=-\lambda\Delta$ and the initial per-position loss is $\lambda^2\|\Delta\|_2^2/d$, a factor $\lambda^2$ larger than for OPRD. Because \cref{eq:rep-loss} is unnormalized, changing $\lambda$ therefore also changes the effective step size. We keep the unnormalized form so that $\lambda=1$ recovers OPRD exactly, and multiply the loss coefficient by $\lambda^{-2}$, which equalizes the initial loss with that of OPRD for every $\lambda$; loss values at different $\lambda$ are compared after this rescaling.
\end{remark}

\paragraph{Diagnostics.}
During training we log the cosine similarity between $h_B$ and $h_T$, the residual norm ratio $\|\Delta\|_2/\|h_T\|_2$ both overall and per layer, and the target norm ratio $\|h^\star\|_2/\|h_T\|_2$, which directly monitors the scale sensitivity described in \cref{rem:scale}.

\end{document}